\documentclass[letterpaper,twocolumn,10pt]{article}
\usepackage{usenix-2020-09}

\usepackage{tikz}
\usepackage{amsmath}
\usepackage{algorithmic}
\usepackage{graphicx}
\usepackage{textcomp}
\usepackage{xcolor}
\usepackage{subfigure}
\usepackage{multirow}
\usepackage{pifont}
\usepackage{threeparttable}
\usepackage{algorithm}

\newcommand{\xmark}{\ding{55}}

\begin{document}

\date{}

\title{\Large \bf FastFold: Reducing AlphaFold Training Time from 11 Days to 67 Hours}


\author{
{\rm Shenggan Cheng\textsuperscript{1}}, {\rm Xuanlei Zhao\textsuperscript{2}}, 
{\rm Guangyang Lu\textsuperscript{2}}, {\rm Jiarui Fang\textsuperscript{2}}, 
{\rm Zhongming Yu\textsuperscript{3}}, \\ 
{\rm Tian Zheng\textsuperscript{4}}, {\rm Ruidong Wu\textsuperscript{5}}, 
{\rm Xiwen Zhang\textsuperscript{5}}, {\rm Jian Peng\textsuperscript{5}}, {\rm Yang You\textsuperscript{1,2}} \\
\textsuperscript{1}National University of Singapore \textsuperscript{2}HPC-AI Technology Inc.\\
\textsuperscript{3}University of California, San Diego \textsuperscript{4}Xi'an Jiaotong University \textsuperscript{5}HeliXon\\
Shenggan Cheng was an employee of HPC-AI Technology Inc. before joining NUS.\\
This work was done by an internship at HPC-AI Technology Inc. (contact@hpcaitech.com)
}

\maketitle

\begin{abstract}

Protein structure prediction helps to understand gene translation and protein function, which is of growing interest and importance in structural biology. The AlphaFold model, which used transformer architecture to achieve atomic-level accuracy in protein structure prediction, was a significant breakthrough. However, training and inference of AlphaFold model are challenging due to its high computation and memory cost. In this work, we present FastFold, an efficient implementation of AlphaFold for both training and inference. We propose \textit{Dynamic Axial Parallelism} and \textit{Duality Async Operations} to improve the scaling efficiency of model parallelism. Besides, AutoChunk is proposed to reduce memory cost by over 80\% during inference by automatically determining the chunk strategy. Experimental results show that FastFold reduces overall training time from 11 days to 67 hours and achieves $7.5\sim9.5\times$ speedup for long-sequence inference. Furthermore, we scale FastFold to 512 GPUs and achieve an aggregate throughput of 6.02 PetaFLOP/s with 90.1\% parallel efficiency.

\end{abstract}

\section{Introduction}

Predicting the three-dimensional structure of a protein from its amino acid sequence, a field known as protein structure prediction, has been a major area of research in structural biology for over 50 years \cite{anfinsen1973principles}. Accurate protein structure prediction has numerous applications, including drug design \cite{thornton2021alphafold} and protein design \cite{jendrusch2021alphadesign}, and can be achieved through both experimental and computational methods. However, experimental methods can be difficult and costly, making computational approaches an attractive option due to their ability to predict protein structure at high throughput and low cost. Improving the efficiency and accuracy of computational protein structure prediction methods is therefore of great importance.

The success of deep neural networks in various fields, such as Computer Vision (CV) and Natural Language Processing (NLP), has led to the widespread use of Artificial Intelligence in many domains. In protein structure prediction, Convolutional Neural Networks (CNNs) were introduced by AlphaFold \cite{senior2020improved} and RaptorX-Contact \cite{xu2019distance} and achieved significant performance improvements. This demonstrates that CNNs can be an effective solution for protein structure prediction using deep learning techniques.

The Transformer model, which uses Multi-Head Attention to focus on different positions and capture long-range dependencies in long sequences \cite{vaswani2017attention}, has made significant improvements in the fields of NLP and CV, and has become the dominant model architecture for tasks such as BERT \cite{devlin2018bert}, GPT \cite{brown2020language}, and ViT \cite{ dosovitskiy2020image}. AlphaFold 2 \cite{jumper2021highly} was the first model to apply Transformer to protein structure prediction and achieved atomic resolution. For the remainder of this paper, we will refer to the transformer-based AlphaFold 2 model as simply AlphaFold.

Although Transformer delivers impressive performance in prediction accuracy, it poses significant computational challenges for training and inference. Firstly, AlphaFold's computational complexity is much higher than that of vanilla Transformer due to its two-dimensional intermediate representation. Secondly, AlphaFold is less computationally efficient on the GPU platform due to its unique model architecture (see Section \ref{section_three}). Thirdly, the limited global batch size prevents training from scaling to more nodes using data parallelism, as larger batch sizes result in a decrease in accuracy. Training AlphaFold on 128 Google TPUv3 nodes takes approximately 11 days \cite{jouppi2017datacenter}. Finally, the high memory consumption of AlphaFold exceeds the capacity of current GPUs. During inference, longer sequences require significantly more GPU memory and can take several hours to process with AlphaFold.

To address these challenges, we introduce FastFold, an efficient implementation of AlphaFold for training and inference. FastFold includes several innovations, such as \textit{Dynamic Axial Parallelism}, a model parallelism strategy that outperforms existing techniques, and \textit{Duality Async Operations}, a communication optimization method implemented as an extension to PyTorch \cite{paszke2019pytorch}. We also apply low-level optimizations to Evoformer, including \textit{AutoChunk} and kernel optimization, which significantly reduce the demand on GPU memory for long sequence inference and the overall cost of training and inference. To the best of our knowledge, FastFold is the first attempt to optimize the performance of training and inference for protein structure prediction models. FastFold introduced large model training techniques and significantly reduces the time and economic cost of training and inference for the AlphaFold model.

In summary, we make the following contributions:

\begin{itemize}
  \item We proposed \textit{Dynamic Axial Parallelism}, a model parallelism strategy with low communication overhead, and \textit{Duality Async Operations}, a communication optimization method that enables computation-communication overlap.
  \item We implemented a series of low-level optimizations. These included \textit{AutoChunk}, a technique that reduces memory usage by over 80\% through automatic chunking, and computational optimization, which resulted in significant speedups. 
  \item We successfully scaled the AlphaFold training to 512 NVIDIA A100 GPUs and resulting in a total of 6.02 PetaFLOPs. This reduces the overall training time from 11 days to 67 hours and leads to significant cost savings. FastFold also achieved a speedup of 7.5-9.5x for long sequence inference and can handle extremely long sequences.
\end{itemize}

\section{Background}
\label{section_two}

\subsection{Transformer}

\begin{figure}[hbt]
    \centering
    \includegraphics[width=0.40\textwidth]{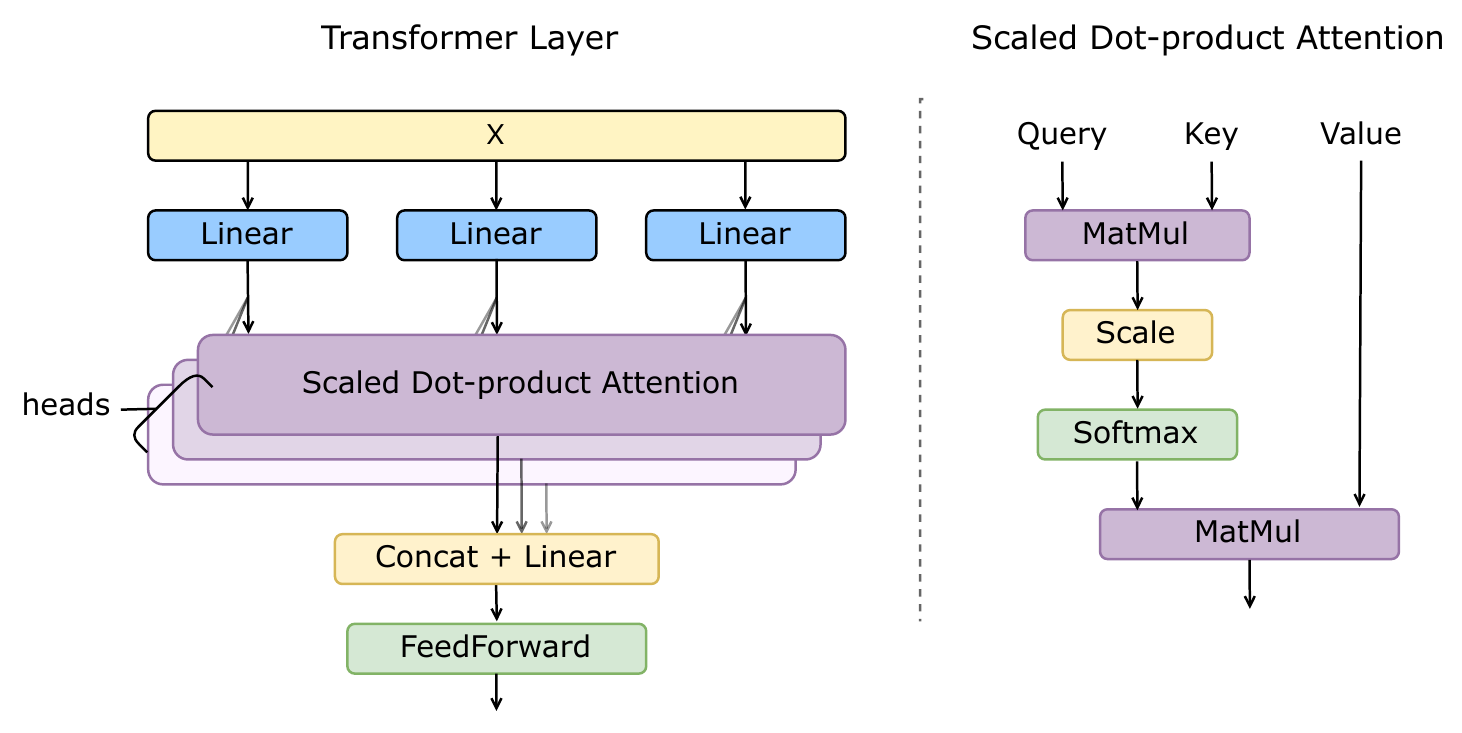}
    \vspace{-5pt}
    \caption{Architecture of the rransformer layer.}
    \label{fig:mha}
\end{figure}

Instead of using Recurrent Neural Networks (RNNs) or Convolutional Neural Networks (CNNs), the transformer models use stacks of self-attention layers to process input with variable sequence lengths. The architecture of transformer layer is shown in Figure \ref{fig:mha}.

The transformer layer consists of two parts: a multi-head attention (MHA) block and a feed-forward block. MHA block is the primary component of the transformer, responsible for modeling the sequence. It is composed of three parts: QKV linear layers divided into heads, scaled dot-product attention, and output linear layer. The input sequence is passed through the three linear layers to obtain \textit{Query}, \textit{Key}, \textit{Value}, which are then split into multiple heads. In the scaled dot-product attention, the dot products of the queries and keys are first calculated, followed by the softmax function, and finally a matrix multiplication with the values. The attention output from each head is concatenated and passed through an output linear layer. The feed-forward block consists of two linear layers and helps to increase the model's capacity.

In the transformer model, two important parameters are the hidden dimension and the number of heads. The hidden dimension refers to the number of features in the input sequences, and the number of heads refers to the number of heads in the MHA.

\subsection{Overview of AlphaFold}

AlphaFold is a transformer-based model that takes amino acid sequences as input and directly outputs the three-dimensional structure of proteins. It obtains Multiple Sequence Alignment (MSA) and templates information for the target sequence through genetic database search and structure database search. MSA information consists of amino acid sequences that are similar to the target sequence, and allows for the identification of amino acids that have mutated during evolution. These co-evolving residues are likely to be located near each other in the three-dimensional structure of the protein. Templates information provides structural information for known sequences, which helps in the prediction of protein structure.

\begin{figure*}[hbt]
    \centering
    \includegraphics[width=0.83\textwidth]{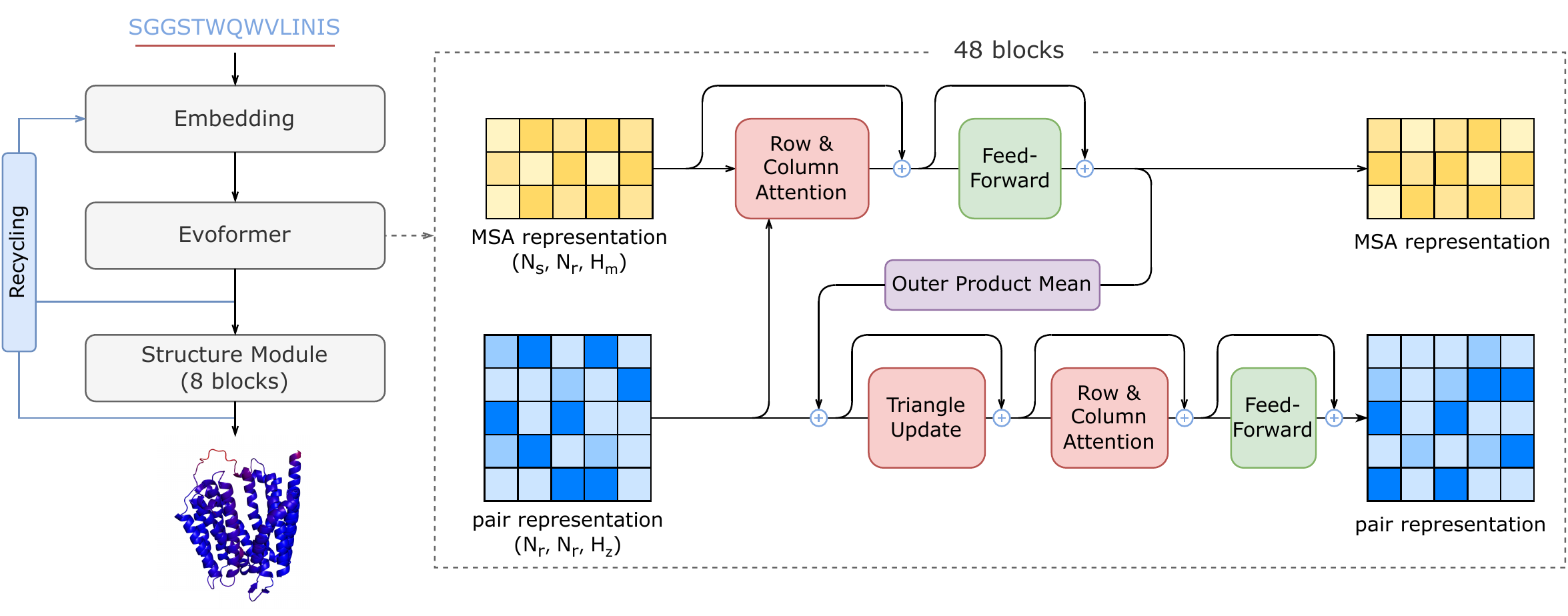}
    \vspace{-5pt}
    \caption{The architecture of AlphaFold model. The amino acid sequence is encoded into MSA and pair representation after Embedding layer, then feeding into Evoformer and Structure Module. In Evoformer, MSA and pair representation were processed by \textit{MSA stack} and \textit{pair stack}, respectively. The number of residues in the input is denoted as $N_r$, while the number of sequences processed in the MSA stack is denoted as $N_s$. The hidden dimensions for MSA and pair representation are set to ($N_m=256$) and ($N_z=128$), respectively.}
    \label{fig:alphafold_arch}
\end{figure*}

The architecture of AlphaFold is shown in Figure \ref{fig:alphafold_arch} and consists of three parts: embedding, evoformer, and structure module. The embedding part encodes the MSA and template information of the target sequence into MSA and pair representations, which are then processed by evoformer blocks (to be discussed in the next section). The MSA and pair representations, which contain highly processed modeling information, are then fed into the structure module, which ultimately outputs the three-dimensional structure.

To reduce training time and memory consumption, AlphaFold uses Bfloat16 precision \cite{kalamkar2019study}. To improve prediction accuracy, AlphaFold uses a recycling technique that repeatedly performs forward passes on the model by re-embedding its output back into the representation. This allows the model to process multiple versions of the embedding features. During training, the number of recyclings is randomly selected between 1 and 4, while it is fixed at 4 during inference.

The training process of AlphaFold consists of \textit{initial training} and \textit{fine-tuning}, as shown in Table \ref{tab:alphafold_training}. It is conducted on 128 TPUv3 cores with a mini-batch size of 128. To ensure the accuracy of the final model, batch size does not exceed 128 during training. The limited batch size prevents AlphaFold from scaling to more devices, leading to an overall training time of 11 days.

\begin{table}[htb]
\small
\centering
\renewcommand\arraystretch{1.1}
\caption{Details of AlphaFold model training.}
\label{tab:alphafold_training}
\begin{tabular}{lcc}
\noalign{\hrule height 1pt}
Model                                     & Initial Training & Fine-tuning  \\ \hline
Residues sequence $N_r$                        & 256              & 384          \\
Number of sequences $N_s$                      & 128              & 512          \\
Batch size                                    & 128 & 128          \\
Training samples ($\times 10^6$)            & $\approx 10$         & $\approx 1.5$    \\
Training time                             & $\approx 7$ days     & $\approx 4$ days \\ \noalign{\hrule height 1pt}
\end{tabular}
\end{table}

\subsection{Evoformer}

The main network trunk of AlphaFold consists of 48 evoformer blocks, each of which has three parts: \textit{MSA stack}, \textit{communication}, and \textit{pair stack}, as shown on the right side of Figure \ref{fig:alphafold_arch}. Evoformer takes two inputs: MSA and template representations, which have two sequence dimensions, unlike the inputs of vanilla transformer. Attention is calculated along different dimensions, which can be divided into row-wise and column-wise.

\begin{figure}[hbt]
    \centering
    \includegraphics[width=0.44\textwidth]{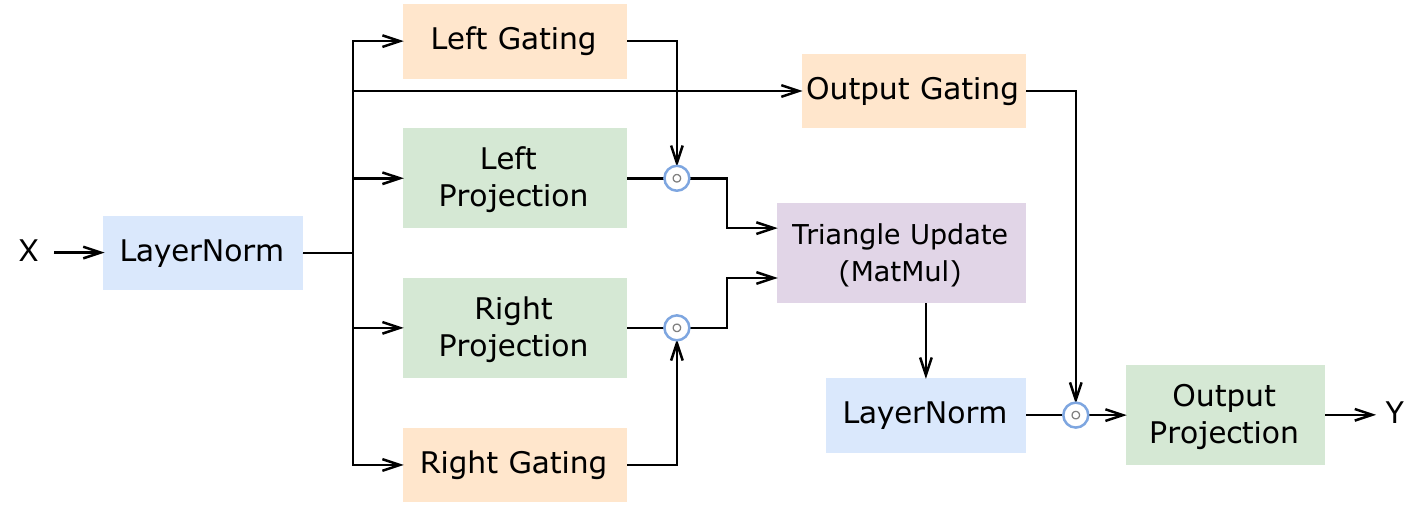}
    \caption{Triangular updates module in evoformer.}
    \label{fig:triangle}
\end{figure}

MSA representations are processed with row-wise attention, column-wise attention, and feed-forward, while pair representations are processed with similar blocks with additional \textit{triangular updates module} (shown in Figure \ref{fig:triangle}). Triangular updates module uses triangular relationships in pair information to infer and update representations. \textit{Attention bias} and \textit{outer product mean} are used to enable communication between the two representations.

\subsection{Parallelism for Training}
\label{para_train}

In modern deep learning training, parallel methods are introduced for two main purposes: 1) to significantly reduce the time cost of training; 2) to train large models with limited resources. The most mainstream parallel methods include data parallelism and model parallelism.

Data parallelism is the most basic and widely used parallel method. Each device has a complete set of model parameters and processes a different mini-batch of training data. During the training phase, each device calculates the local gradient using its own mini-batch, then uses all-reduce communication to average the gradients globally. The model parameters are then updated based on the averaged gradients.

Model parallelism distributes the model parameters across multiple devices, which can be divided into tensor parallelism and pipeline parallelism based on the distribution method. In pipeline parallelism, the model is split vertically (layer-wise) among the devices. However, this method introduces device idleness due to dependencies between the computations on different devices. To improve resource utilization, the mini-batch is often divided into micro-batches, which allows for more overlap between computations on different devices.

Tensor parallelism is typically imposed on the linear layer because it is relatively easy to distribute matrix multiplication across different devices. To reduce communication overhead, Megatron-LM \cite{10.1145/3458817.3476209} proposed column parallelism and row parallelism. In column parallelism, the weight matrix $W$ is divided column-wise across $N$ devices, resulting in $N$ matrices $W_1, W_2, ..., W_n$. Matrix multiplications $XW_1, XW_2, ..., XW_n$ are conducted in parallel, resulting in $N$ output vectors $Y_1, Y_2, ..., Y_n$. In row parallelism, the weight matrix $W$ and input vectors $X$ are divided across $N$ devices, and matrix multiplications $X_1W_1, X_2W_2, ..., X_nW_n$ are conducted in parallel, resulting in $N$ output vectors. The final output vector $Y$ is obtained through an all-reduce operation. In the feed-forward block, column parallelism can be used in the first linear layer and row parallelism in the second layer. In the MHA block, Megatron-LM uses column parallelism for the QKV linear layer and row parallelism for the output linear layer.

\section{In-depth Analysis of Evoformer}
\label{section_three}


For the convenience of later expressions, we denote the number of residues in the input as $N_r$, the number of sequences processed in the \textit{MSA stack} as $N_s$, the hidden dim for MSA representation as $H_m = 256$, the hidden dim for pair representation as $H_z = 128$. Specific values of $N_r$ and $N_s$ can be found in Table \ref{tab:alphafold_training}.

\subsection{Performance Analysis}

To further analyze the performance characteristics of evoformer, we can classify its operators into three main categories based on the characteristics of the computation and memory access:

\vspace{-\topsep}
\begin{itemize}
    \setlength{\parskip}{0pt}
    \setlength{\itemsep}{0pt plus 1pt}
    \item \textit{General Matrix Multiply (GEMM)}: This category includes matrix multiplication, batch matrix-matrix product, and other dense matrix calculations.
    \item \textit{Batch Reduction}: This category includes LayerNorm, Softmax, and other operations with lower computational intensity.
    \item \textit{Element-wise Operators}: This category includes element-wise addition, dropout, and activations, and is the least compute-intensive category.
\end{itemize}
\vspace{-\topsep}

GEMM operators are typically computed using highly optimized Basic Linear Algebra Subprograms (BLAS) libraries provided by the vendor, such as cuBLAS on GPU platforms and MKL (Math Kernel Library) on CPU platforms. However, deep learning frameworks like PyTorch may not be as efficient at implementing non-GEMM operators. For example, during AlphaFold model training on NVIDIA Tesla A100, only 14.7\% of the time was spent on GEMM operators, while 55.7\% was spent on batch reduction, 19.8\% on element-wise operations, and 9.8\% on other operations such as data movement. The time spent on batch reduction, in particular, was high because the implementation of LayerNorm and Softmax in PyTorch is inefficient. This performance issue also occurs with other Transformer models, but is more severe in AlphaFold due to its smaller hidden dimension (as shown in Table \ref{tab:evoformer_transformer}). PyTorch's native batch reduction kernel is less efficient with small hidden dimensions.

\begin{table}[htb]
\small
\centering
\renewcommand\arraystretch{1.2}
\caption{Different setting of evoformer in AlphaFold and transformer in ViT and GPT.}
\vspace{5pt}
\label{tab:evoformer_transformer}
\begin{tabular}{cccc}
\noalign{\hrule height 1pt}
                     & AlphaFold  & ViT-B/16 & GPT-2       \\ \hline
Sequence Shape       & ($N_s$, $N_r$) or ($N_r$, $N_r$) & 196      & 512 \\
Layers                & 48         & 12       & 48          \\
Hidden Dim           & 128 or 256 & 768      & 1600        \\
Heads                 & 8 or 4     & 12       & 25          \\
Params / Layer & 1.8 M      & 7.1 M    & 30.7 M      \\ \noalign{\hrule height 1pt}
\end{tabular}
\end{table}

\subsection{Memory Consumption}
\label{memory_analysis}

During AlphaFold training, we observed high memory consumption. It is worth noting that the overall model size for AlphaFold is only 93M, according to Table \ref{tab:evoformer_transformer}. Despite its small model size, AlphaFold requires a large amount of memory due to the large intermediate activations it generates. For example, the activations in the attention module require $N_r^3 \times N_{head} \times sizeof(BFloat16)$ bytes of memory, which can exceed 20 GB for 48 layers when $N_r=384$ and $N_{head}=4$. To mitigate this issue, AlphaFold uses gradient checkpointing \cite{chen2016training} to reduce memory consumption. However, memory is still a bottleneck for AlphaFold, as each device can only process one data sample during training due to the limited memory capacity. 

In the inference stage, long sequences also result in very large tensor buffers, further stressing the memory capacity. AlphaFold uses a chunking technique to reduce peak memory consumption. However, most operations in the evoformer module use less than 20\% of peak memory. The fixed chunking scheme used in AlphaFold, which cannot adapt to the range and size of chunks, makes it difficult to perform long-sequence inference within a fixed memory budget.
\section{Parallel Evoformer}

Introducing model parallelism allows the training to be distributed across more computational resources, reducing the overall training time. For inference, model parallelism can significantly reduce the latency of long-sequence predictions, making it more practical for use in real-world applications.

As we described in the background section, pipeline parallelism requires further partitioning of mini-batch into multiple micro-batches in order to improve the efficiency of hardware resource utilization. However, in the training of AlphaFold, due to the batch size limitation, it is not feasible to partition the mini-batch, so pipeline parallelism is not a suitable parallelism strategy for AlphaFold.

\subsection{\textbf{Tensor Parallelism}}
\label{TP_alphafold}

Unlike pipeline parallelism, Tensor Parallelism (TP) can be easily adapted to AlphaFold model training. The main structure of evoformer contains attention blocks and feed-forward blocks, similar to the structure of vanilla transformer. Therefore, we can use TP in a similar way to Megatron-LM, as described in Section \ref{para_train}.

However, TP is not efficient for AlphaFold for the following reasons: \textbf{1)} frequent synchronization communication in each evoformer layer leads to high overhead; \textbf{2)} modules other than attention and feed-forward cannot be parallelized; \textbf{3)} the scaling of TP is limited by the number of heads in attention (the heads in the AlphaFold are 4 in the pair stack, so TP can be scaled up to a maximum of 4 devices).

\subsection{\textbf{Dynamic Axial Parallelism}}

\begin{figure}[hbt]
  \centering
  \subfigure[Transpose] {
    \label{fig:dap_attention}     
    \includegraphics[width=0.78\columnwidth]{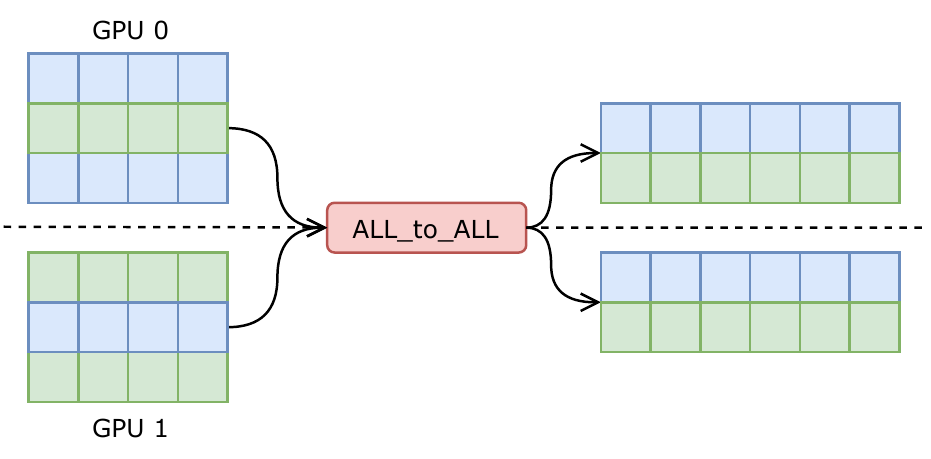}  
  }    
    
  \subfigure[Outer Product Mean] { 
    \label{fig:dap_outer}     
    \includegraphics[width=0.88\columnwidth]{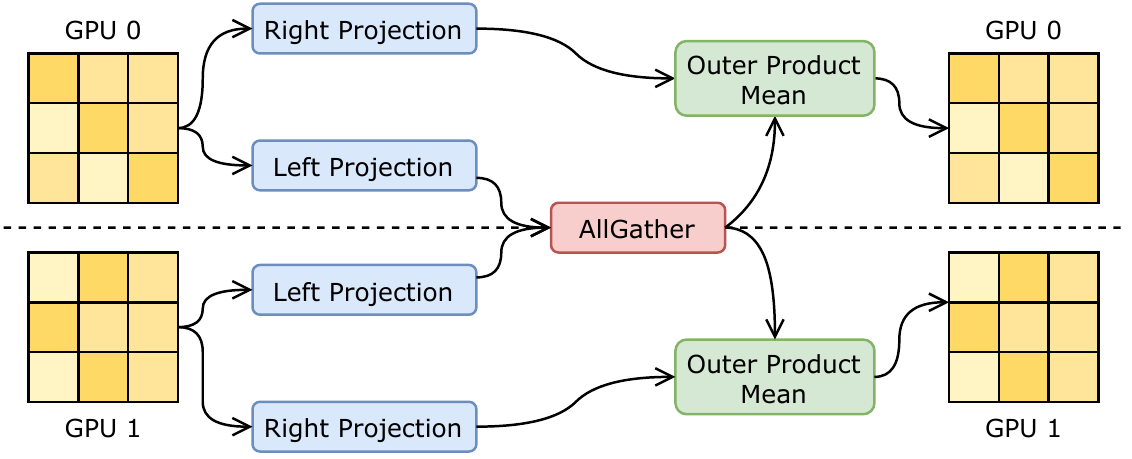}
  }    
  \vspace{-5pt}
  \caption{Communication for Dynamic Axial Parallelism. The all-to-all operation is required during transpose, while the all-gather operation must be inserted in the outer product mean module and triangular updates module.}     
  \label{fig:dap}
\end{figure}

To address these problems, we propose \textit{Dynamic Axial Parallelism} (DAP). In AlphaFold training, the model has relatively few parameters but relatively large activations. Therefore, unlike TP, we choose to keep the complete model parameters on each device and divide the input and activations among different devices. Both MSA representation and pair representation processed by the evoformer module contain two sequence dimensions, but the calculations in the evoformer are along only one dimension. Therefore,  we can divide the data along the other dimension and insert all-to-all communication when the two sequence dimensions are transposed, keeping the data dimensions of the computation axial and complete on each device, as shown in Figure \ref{fig:dap_attention}. No other communication is needed in the computation of attention computation. In the outer product mean module, we need to gather the global left projection using all-gather and then perform the outer product mean with the local right projection. The triangular updates module also uses a similar approach for parallelism.

In Table \ref{tab:tp_dap_compare}, we compare the communication overhead of Tensor Parallelism (TP) and Dynamic Axial Parallelism (DAP). It can be seen that TP only supports parallelism in the attention and feed-forward modules, while DAP supports all the computational modules of the evoformer. TP introduces 12 all-reduce communications in the attention and feed-forward module, 6 in the forward pass and 6 in the backward pass. Assuming the use of a ring all-reduce, the amount of communication per step is calculated as $24 \times K(N-1) / N$, where $K$ is the size of the intermediate representation and $N$ is the number of devices for model parallelism. In the forward pass, DAP introduces one all-gather communication in the outer product mean module and two in the triangular updates modules (incoming and outgoing). The communication volume per all-gather is $(N-1) \times K / N$. The backward pass does not require additional communication. DAP needs to insert all-to-all communication in between calculations in different dimensions, a total of 12 times (6 times in the forward and 6 times in the backward) in an evoformer block. Each all-to-all requires $(N-1) \times K / N^2$ of communication volume. Overall, DAP has an order of magnitude lower communication volume compared to TP. 

\begin{table}[hbt]
\small
\centering
\renewcommand\arraystretch{1.3}
\begin{threeparttable}
\caption{Communication Volume for Each Evoformer Block.}
\label{tab:tp_dap_compare}
\begin{tabular}{lcc}
\noalign{\hrule height 1pt}
                  & TP & DAP \\ \hline
Attention+FF       & $24 \times K (N-1) / N $  & No Comm          \\
Outer Product Mean & \xmark         & $K (N-1) / N$              \\
Triangle Update    & \xmark         & $2 \times K (N-1) / N$               \\
Transpose          & No Comm        & $12 \times K (N-1) / N^2 $            \\ \noalign{\hrule height 1pt}
\end{tabular}
\begin{tablenotes}
  \item $K$: The size of the intermediate activation.
  \item $N$: The number of devices for model parallelism.
\end{tablenotes}
\end{threeparttable}
\vspace{-10pt}
\end{table}

Therefore, DAP has several advantages over TP: \textbf{1)} DAP supports all computational modules in Evoformer; \textbf{2)} The communication volume of DAP is much smaller than TP; \textbf{3)} Model parallelism can distribute activation to different devices, and DAP consumes less memory than TP because it has more parallel parts; \textbf{4)}  DAP has more opportunities for communication optimization, such as computation-communication overlap.

\section{Low-level Optimization}

Dynamic Axial Parallelism (DAP) enables the training and inference of AlphaFold to be distributed over more computational resources, significantly reducing the time cost. To further optimize the time and economic cost, we implement several low-level optimization techniques, including communication and computation optimization. We also propose AutoChunk, a method that automatically determines the optimal chunking strategy to efficiently inference long sequences with minimal computational overhead.

\subsection{Communication Optimization}

DAP requires all-to-all and all-gather communication between all devices with axial parallelism. Due to the synchronized communication in the layer, the communication of DAP can become a bottleneck. To address this issue, we design and implemente optimization strategies to reduce the communication overhead of DAP.

In PyTorch, all computation and communication are assigned to different CUDA streams. However, PyTorch will block the computation stream to wait for the completion of the communication. In the vanilla transformer model, the computation is straightforward and there is no opportunity to overlap the communication with computation. However, in AlphaFold, we have the opportunity to overlap the computation and communication because we have two representation features to process. While it is difficult to use asynchronous communication interfaces and implement corresponding communication in the backward pass in dynamic-graph deep learning frameworks like PyTorch, we have designed the \textit{Duality Async Operations (DAO)} for PyTorch to enable the overlap of communication and computation.

\begin{figure}[hbt]
    \centering
    \includegraphics[width=0.44\textwidth]{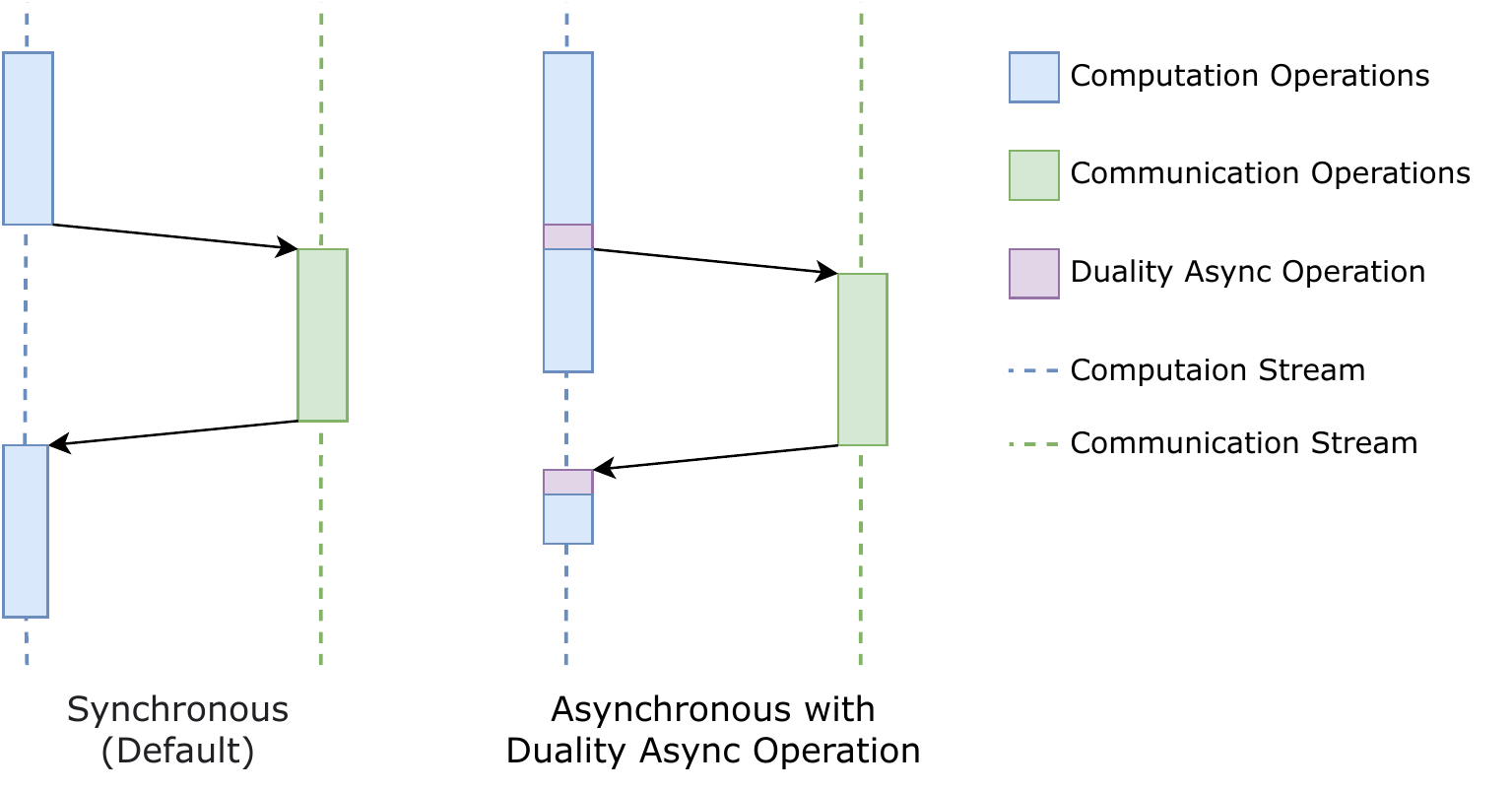}
    \caption{The \textit{Duality Async Operations (DAO)} enables the overlap of computation and communication in both the forward and backward passes through asynchronous execution.}
    \label{fig:async_op}
\end{figure}

As shown in Figure \ref{fig:async_op}, the "Duality Async Operation" consists of a pair of communication operations. During the forward pass, the first operation triggers asynchronous communication, followed by computation on the computation stream that does not depend on the communication. The second operation then blocks the asynchronous communication until it is completed, after which the subsequent computation is performed. In the backward pass, the second operation triggers the asynchronous communication and the first operation blocks it. We have observed that using asynchronous communication significantly reduces the communication overhead through the overlap of computation and communication.

We implement the DAO using the PyTorch Autogard Function, which is part of PyTorch's automatic differentiation package. The PyTorch Autogard Function allows us to define the forward and backward computations independently. We enable inter-operation between these two operations by passing asynchronous communication requests between two operators, thus allowing us to trigger and block asynchronous communication as needed.

\subsection{Computation Optimization}

As previously mentioned, the GEMM operators in AlphaFold only account for a small portion of the total runtime, indicating that there is significant potential for improving performance in the non-GEMM part. To achieve high performance, we implemented several optimization techniques, including kernel fusion and highly optimized kernels for batch reduction operations (softmax and layernorm).

The softmax function is a normalized exponential function that converts its input elements into values between 0 and 1, with the sum of all elements being 1. In the AlphaFold model, the input to the softmax function has many rows, but each row has a relatively small number of elements. If not implemented and parallelized properly, the native kernel will have poor performance in this case. The input to the softmax function goes through two additions - one due to the mask and one due to the bias\_add operation in the evoformer's attention mechanism. These broadcast additions introduce a significant memory bottleneck.

For small column sizes, we use one warp to calculate one row of data and use the communication primitives between registers to achieve high performance. To calculate the global max of a row, we first find the local max in threads and then use WarpAllReduce to get the global max. Subtraction and exponential operations are performed, and the local sum is calculated in threads. The global sum is then obtained using WarpAllReduce, followed by a final division. In addition, we have fused the mask and bias\_add into the softmax kernel, thereby avoiding broadcasts and significantly improving performance.

We also used the same approach to implement a high-performance layernorm kernel. To further improve the efficiency of the computation, we applied other kernel fusion techniques to reduce memory access overhead and lower kernel startup overhead. We implemented kernel fusion in two approaches in the AlphaFold model:

\vspace{-\topsep}
\begin{itemize}
    \setlength{\parskip}{0pt}
    \setlength{\itemsep}{0pt plus 1pt}
    \item \textit{Merge GEMM.} In the attention computation of query, key, and value, we merge three linear layers into one. In the Triangular Updates Module, we also merge the left project with the right project and the left gating with the right gating.
    \item \textit{JIT Fusion.} PyTorch's Just-In-Time (JIT) compilation can generate optimized fused kernels. We combine several element-wise operations (\textit{bias + sigmod + element-wise product} and \textit{bias + dropout + add}) into fused kernels by PyTorch JIT.
\end{itemize}
\vspace{-\topsep}

\subsection{AutoChunk}

High memory consumption is a major bottleneck in the AlphaFold model, especially for long sequences inference. To address this issue, AlphaFold uses chunk techniques, which involve partitioning the tensor along dimensions that are independent of the computation. This technique has been demonstrated to significantly reduce the peak memory consumption, such as the attention module. However, this approach has several drawbacks: \textbf{1)} it requires significant programming effort from expert technicians; \textbf{2)} manually analyzing and specifying of the range and size of chunk can be labor-intensive; \textbf{3)} human-designed chunk schemes can be inefficient, resulting in increase inference latency. Our analysis also shows that 95\% of the operations in evoformer have a memory footprint below 20\% of the peak, suggesting that module-level chunking may not be necessary. Instead, targeting and optimizing these outlying operations may provide an efficient means of chunking.

Based on this observation, we propose AutoChunk, which is able to generate chunk strategies adaptively and efficiently for inference. AutoChunk can identify the optimal chunk range and chunk size, reducing memory usage with minimal cost. The overview of AutoChunk is shown in Algorithm \ref{alg:autochunk}. Given the computational graph $G$ and memory budget $M$ as inputs, AutoChunk iteratively finds all chunk strategies $C$. In each iteration, it estimates the memory consumption based on the existing chunk strategy $C$ and graph $G$, and finds the node with the highest memory usage, where node refers to a basic operation such as add and linear. Then, it determines the maximum chunk range according to the peak memory node and current memory status, and identifies all possible chunk strategies within that range. Finally, the best strategy is selected and added to the chunk strategy $C$. Once all chunk strategies $C$ have been found, they are inserted into the code through code generation.

\begin{algorithm} 
    \caption{AutoChunk} 
    \label{alg1}
    \renewcommand{\algorithmicrequire}{\textbf{Input:}}
    \renewcommand{\algorithmicensure}{\textbf{Output:}}
    \label{alg:autochunk} 
    \begin{algorithmic}
	\REQUIRE Computational graph $G$ and memory budget $M$ 
   	\ENSURE Code with chunk $O$
        \STATE Initialization: Chunks $ C \gets \emptyset$
	\WHILE{$best\_chunk \neq None$} 
	\STATE $peak\_mem \gets $ EstimateMemory($G,C$)
        \STATE $max\_chunk \gets $ FindMaxChunk($G,C,peak\_mem$)
        \STATE $pos\_chunk \gets $ FindPossibleChunk($G,C,max\_chunk$)
        \STATE $best\_chunk \gets $ FindBestChunk($G,C,M,pos\_chunk$)
        \STATE $C \gets C \cup \{best\_chunk\}$
	\ENDWHILE
        \STATE $O \gets $ CodeGen($G,C$)
    \end{algorithmic} 
\end{algorithm}

\begin{figure}[hbt]
    \centering
    \includegraphics[width=0.40\textwidth]{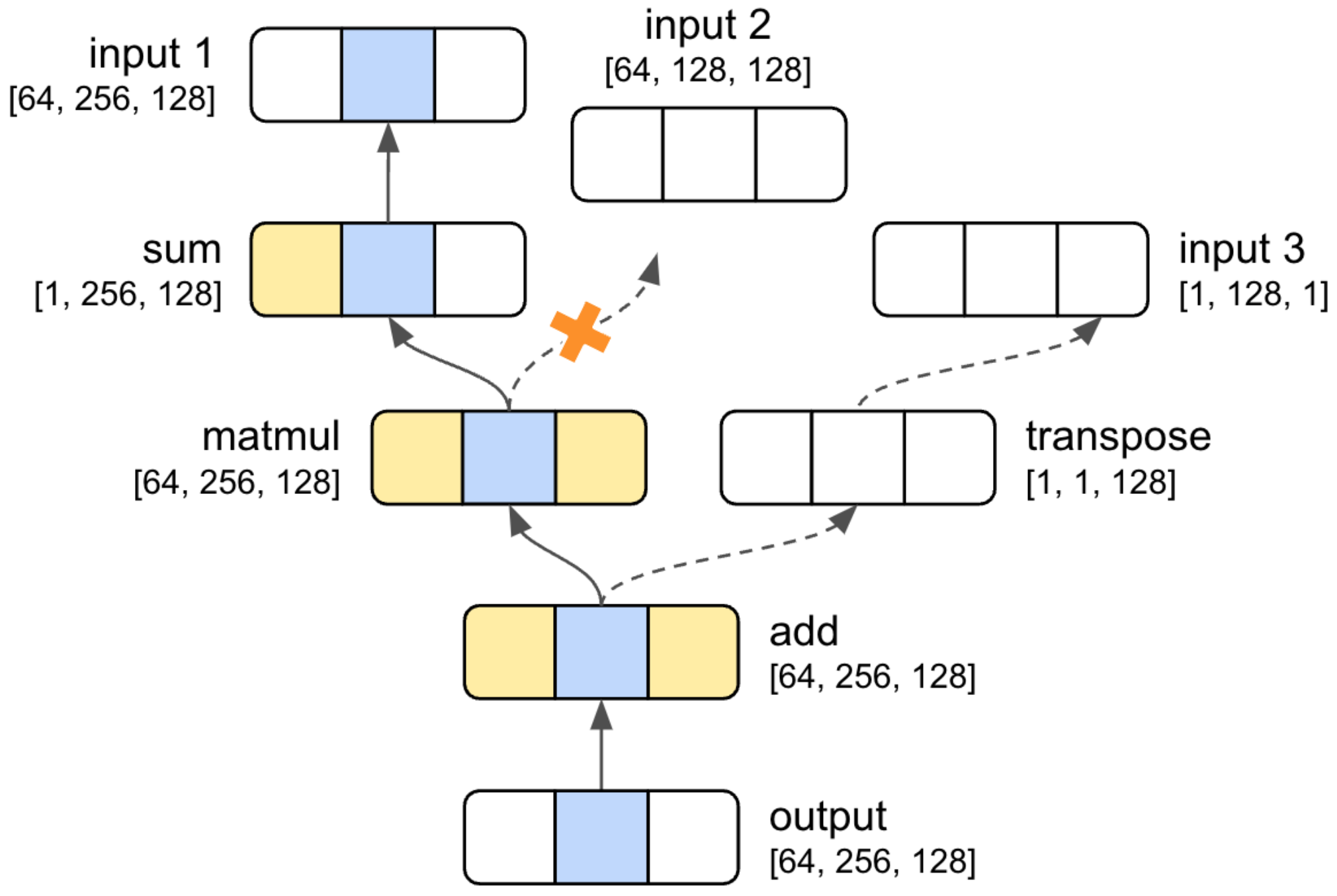}
    \caption{Illustration of AutoChunk strategy search. The boxes represent the dimensions of tensors. Blue boxes are chunk dimensions, and yellow boxes are compute dimensions. Arrows indicate the flow path of chunk dimensions.}
    \label{fig:autochunk}
\end{figure}

Specifically, in order to find the max chunk range, we identify the peak memory node from memory estimation and extend the chunk range from this node. The nature of chunk is partitioning the intermediate memory of the current node and the activation memory of active nodes into smaller parts, where active nodes refer to all nodes currently generated but not deleted in $G$. So the number and size of active nodes are taken into account to determine the max range.

Then we need to define chunk range and identify all possible chunk strategies within this range. Consider a function denoted as:

\[ Y = F(X) \]

Then the chunked function can be denoted is:

\[X^{chunk}_i = [x^1_{c(x^1),i},...,x^{Nx}_{c(x^{Nx}),i}],\]
\[Y^{chunk}_i = [y^1_{c(y^1),i},...,y^{Ny}_{c(y^{Ny}),i}],\]
\[ Y^{chunk}_i = F(X^{chunk}_i,X^{nonchunk})\]

Where $X^{chunk}$ are chunked inputs, $X^{nonchunk}$ are inputs without chunk, and $x^k_{c(x^k),i}$ refers to the $i$-th part of $x^k$ partitioned by the chunk dimension $c(x^k)$.

As seen in the equation above, chunk involves identifying suitable function $F$ and partitioning its inputs and outputs along certain dimensions, and then combining the results to obtain the output. From this definition of chunking, we can define the criteria for a reasonable chunk range as follows: \textbf{1)} All outputs $Y=[y^1,...,y^{Ny}]$ have a chunkable dimension $c(y^j)$. \textbf{2)} For all nodes $N=[n^1, ..., n^{Nn}]$ within the chunk range, if any dimension $d^{n^k}_l$ of a node $n^k$ belongs to $S(c(Y))$, where $S(c(Y))$ refers to the dimensions included in the flow path of outputs chunk dimension $c(Y)$, then dimension $d^{n^k}_l$ needs to be chunked, i.e. $c(n^k)=d^{n^k}_l$. \textbf{3)} The chunked dimension $c(n^k)$ of each node must be a free dimension, without involving any computation. A simple illustration is shown in Figure \ref{fig:autochunk}. Search begins from the chunk dimension of output, and trace upwards as the arrow indicates. Dimensions that trace go through are denoted as chunk dimensions.

As defined above, we need to trace each output upwards for every chunk range, which is computationally expensive. Therefore, we propose an efficient two-stage search method: in the first stage, we only consider whether the start and end nodes of the chunk range satisfy rules 2 and 3 mentioned above. In the second stage, we further search within the satisfied range to check if all nodes within the chunk range meet the criterion for a possible chunk.

Subsequently, we select the optimal chunk strategy and determine the chunk size. We aim to select the chunk with the least impact on the speed within the memory budget. Some nodes are rearranged to optimize chunk efficiency, i.e. transpose node can be removed from chunk in Figure \ref{fig:autochunk}. Once we have obtained all the chunk ranges, we utilize them for code generation.

\section{Evaluation}

In this section, we will evaluate the end-to-end improvements provided by FastFold in both training and inference, and compare the performance of DAP to TP. And then we analyze the enhancements contributed by FastFold's low-level optimizations, including computation optimizations and AutoChunk. All experiments were conducted on the NVIDIA Tesla A100 platform, using the official implementation of AlphaFold and OpenFold \cite{Ahdritz_OpenFold_2021} as baselines. The official implementation of AlphaFold includes only the inference part, while OpenFold reproduces both training and inference based on the original AlphaFold paper \cite{jumper2021highly}."

\subsection{End-to-End Training Performance}

\begin{figure*}[htb]
  \centering
  \subfigure[Intra-node Parallelism (Initial Training)] {
    \centering
    \label{fig:mp_scale}     
    \includegraphics[width=0.63\columnwidth]{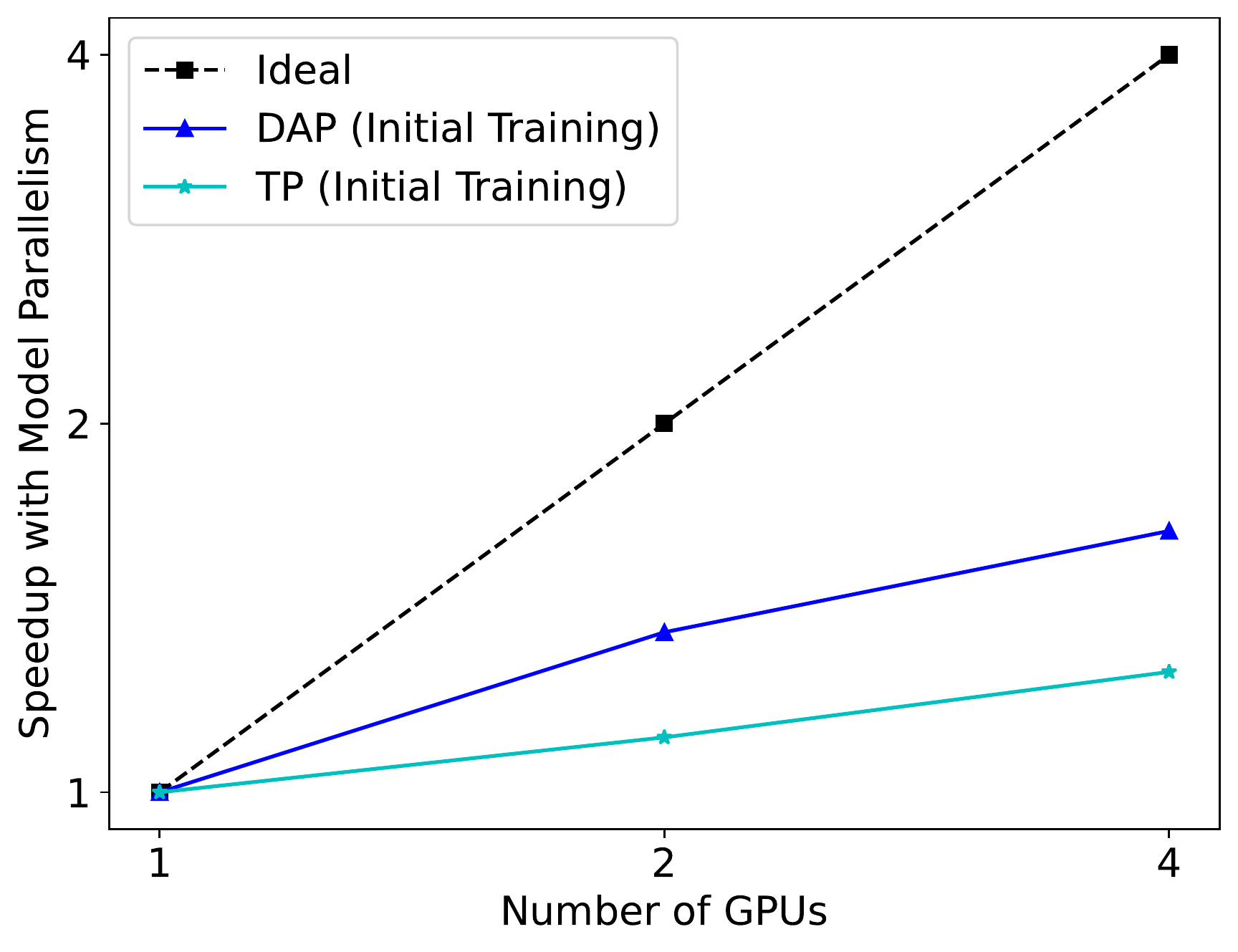}  
  }    
  \hfill
  \subfigure[Intra-node Parallelism (Finetuning)] { 
    \centering
    \label{fig:mp_scale_2}     
    \includegraphics[width=0.63\columnwidth]{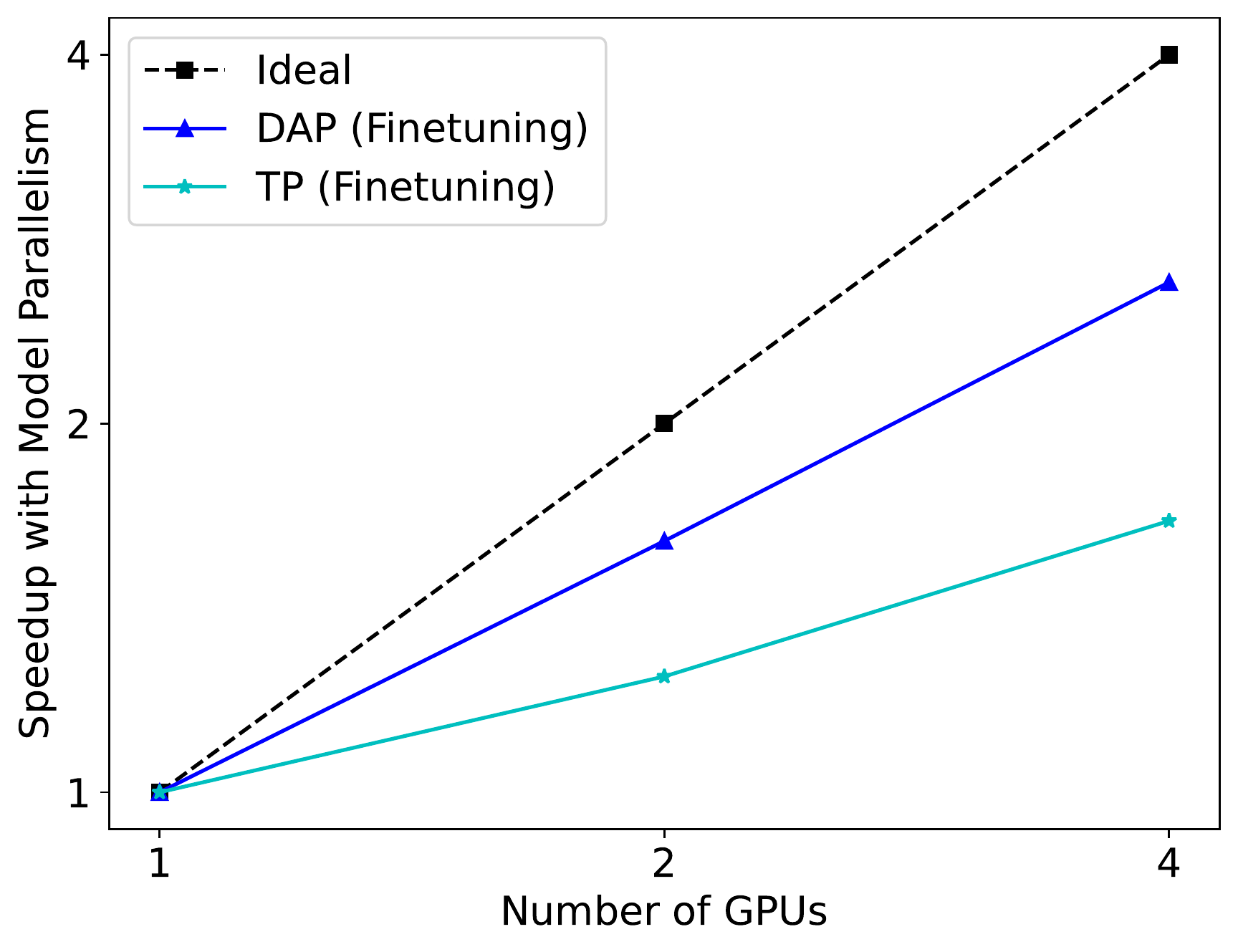}
  }
  \hfill
  \subfigure[Inter-node Parallelism] {
    \centering
    \label{fig:dp_scale}     
    \includegraphics[width=0.65\columnwidth]{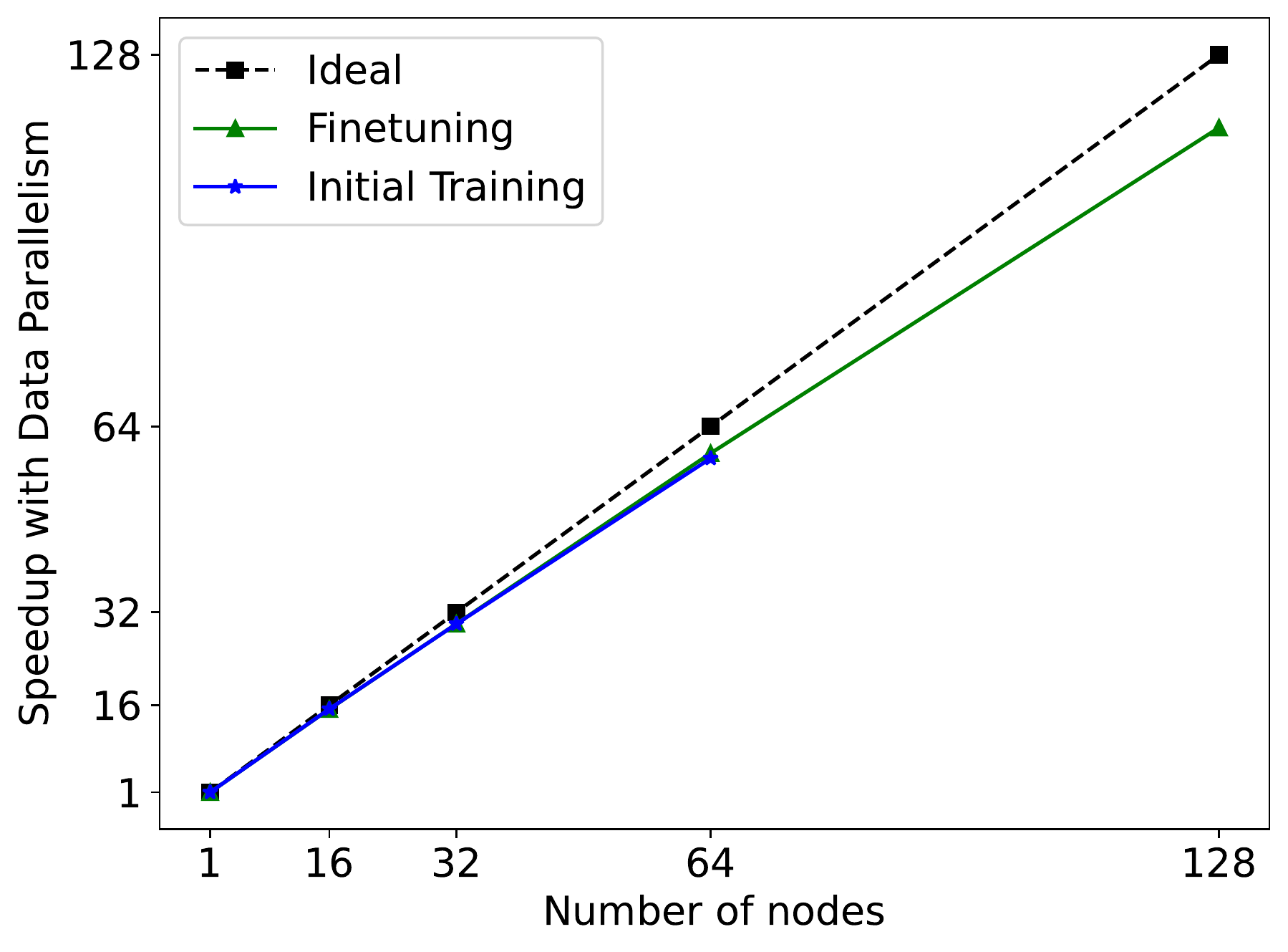}
  }
  \vspace{-5pt}
  \caption{Parallel Efficiency of AlphaFold Training. The left two figures show the scaling efficiency of model parallelism intra-node, and the right figure shows the scaling efficiency of data parallelism inter-node.}     
  \label{fig:parallel_efficiency}
\end{figure*}

In the evaluation of end-to-end training performance, we use the training parameters from the official AlphaFold paper for testing to better compare different methods and implementations in real training scenarios. All training experiments were conducted on a 128-node GPU supercomputer, where each node consists of 4 NVIDIA Tesla A100s and has NVLink for GPU interconnects. Tensor parallelism relies heavily on fast interconnections between devices for communication, so model parallelism is generally used within nodes and data parallelism between nodes during training. We evaluate the training performance at both the model parallelism and data parallelism levels and present the results in Figure \ref{fig:parallel_efficiency}.

\textbf{1) Model Parallelism.} We compared the scalability of TP and DAP for model parallelism under two training settings: Initial Training and Fine-tuning. As shown in Figure \ref{fig:mp_scale} and \ref{fig:mp_scale_2}, DAP has significantly better scalability than TP for both Initial Training and Fine-tuning. The scalability of Initial Training is worse because the sequence length is shorter, which results in more obvious communication overhead. It is worth noting that when Initial Training is scaled to 4 GPUs, we can turn off the activation checkpoint because the GPU memory is sufficient, leading to a 16.14\% performance improvement. The \textit{Duality Async Operation} also significantly reduces the communication overhead, resulting in an overall performance improvement of 3\% to 8\%.

\textbf{2) Data parallelism.} We used data parallelism to scale with fixed model parallelism settings. Following the settings of the official AlphaFold paper, we scaled the global batch size to 128 for data parallelism. In fine-tuning training, we used DAP to scale the computation of a sample to a full node (4 GPUs), so data parallelism was scaled from 1 to 128 nodes. In initial training, to improve scaling efficiency, we only scaled DAP to half a node (2 GPUs), so data parallelism was scaled to 64 nodes only. The scaling results are shown in Figure \ref{fig:dp_scale}. It can be seen that data parallelism scales almost linearly and the scaling efficiency of Fine-tuning training reaches 90.1\%.

\begin{table*}[htb]
\small
\centering
\renewcommand\arraystretch{1.2}
\caption{Comparison of Resource and Time Cost of Different Implementation.}
\label{tab:overview}
\begin{tabular}{ccccccc}
\noalign{\hrule height 1pt}
Implementation             & Framework                & Training Process & Hardware                     & Step Time  & Training Time  & Resource                          \\ \hline
\multirow{2}{*}{AlphaFold} & \multirow{2}{*}{JAX\cite{frostig2018compiling}}     & Initial training & \multirow{2}{*}{128 $\times$ TPUv3} & /             & \multirow{2}{*}{11 days}   & \multirow{2}{*}{33792  TPU hours} \\
                           &                          & Fine-tuning      &                              & /             &                       &                                   \\ \hline
\multirow{2}{*}{OpenFold}  & \multirow{2}{*}{PyTorch} & Initial training & \multirow{2}{*}{128 $\times$ A100}  & 6.186 s         & \multirow{2}{*}{8.39 days} & \multirow{2}{*}{25774 GPU hours}  \\
                           &                          & Fine-tuning      &                              & 20.657 s       &                       &                                   \\ \hline
\multirow{2}{*}{\textbf{FastFold}}  & \multirow{2}{*}{PyTorch} & Initial training & 256 $\times$ A100                   & 2.487 s        & \multirow{2}{*}{\textbf{2.81} days} & \multirow{2}{*}{\textbf{20738 GPU hours}}  \\
                           &                          & Fine-tuning      & 512 $\times$ A100                   & 4.153 s      &                       &                                   \\ \noalign{\hrule height 1pt}
\end{tabular}
\end{table*}

Table \ref{tab:overview} compares the time and economic costs of three implementations of AlphaFold: OpenFold, FastFold, and the original AlphaFold. These costs are based on the results of our evaluations of training performance. To minimize time and economic cost, we choose to use 256 A100 GPUs for Initial Training and then scale to 512 A100 GPUs during the Fine-tuning phase. With this configuration, FastFold reduces the training time to 2.81 days. This represents a 3.91-fold reduction in training time compared to the original AlphaFold and a 2.98-fold reduction compared to OpenFold, as well as a 20\% reduction in economic cost. During the Fine-tuning phase, FastFold achieve an aggregate throughput of 6.02 PetaFLOP/s with 512 A100 GPUs. These significant reductions in time and economic cost make FastFold a faster and more cost-effective option for training protein structure prediction models, which will drive the efficiency of research and development of related models and facilitate the development of Evoformer-based protein structure prediction models.

\subsection{End-to-End Inference Performance}

We evaluate the inference performance of FastFold, OpenFold, and AlphaFold implementations in three scenarios: short sequences, long sequences, and extremely long sequences. The experiments are conducted on a GPU server with 8 NVIDIA A100s (with NVLink). In practice, it is common to use multiple models and aggregate their results to improve the accuracy of protein structure predictions. However, since the performance characteristics of multiple models are consistent, our experiments on inference performance only evaluate the performance of a single model. Additionally, some optional modules such as ExtraMSA and templates are disabled.

\begin{figure}[hbt]
    \centering
    \includegraphics[width=0.41\textwidth]{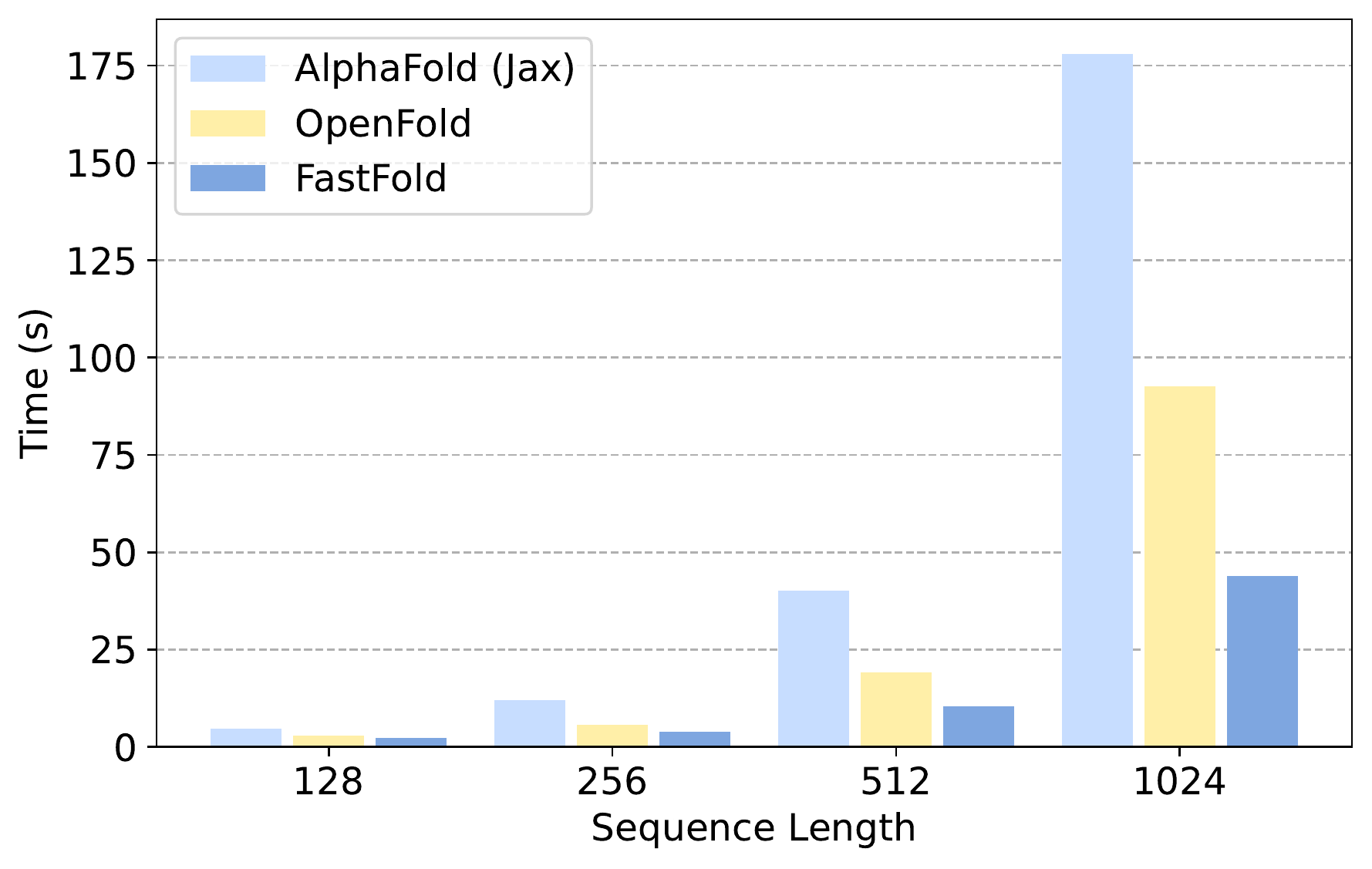}
    \vspace{-8pt}
    \caption{Inference latency for short sequences.}
    \label{fig:inference_small}
\end{figure}

For short sequences, which typically have amino acid sequences no longer than 1K, inference of a single model takes a few seconds to about one minute. At this sequence range, the memory consumption is relatively small and the efficiency of using distributed inference is lower. Therefore, we compared the inference latency of the three implementations on a single GPU and presented the results in Figure \ref{fig:inference_small}. In the scenario of short sequence inference, FastFold's inference performance is improved by 2.01-4.05 times and 1.25-2.11 times compared to AlphaFold and OpenFold, respectively. It is worth noting that AlphaFold's performance is lower on the GPU platform. This is because AlphaFold uses the JAX framework, which has better support for Google TPUs and may not have optimal computational performance on the GPU platform. In addition to the inference time, AlphaFold also requires 50-150 seconds to compile kernels during inference.

\begin{figure*}[htb]
  \centering
  \subfigure[Sequence Length = 1536] {
    \centering
    \label{fig:inference_long_1536}     
    \includegraphics[width=0.60\columnwidth]{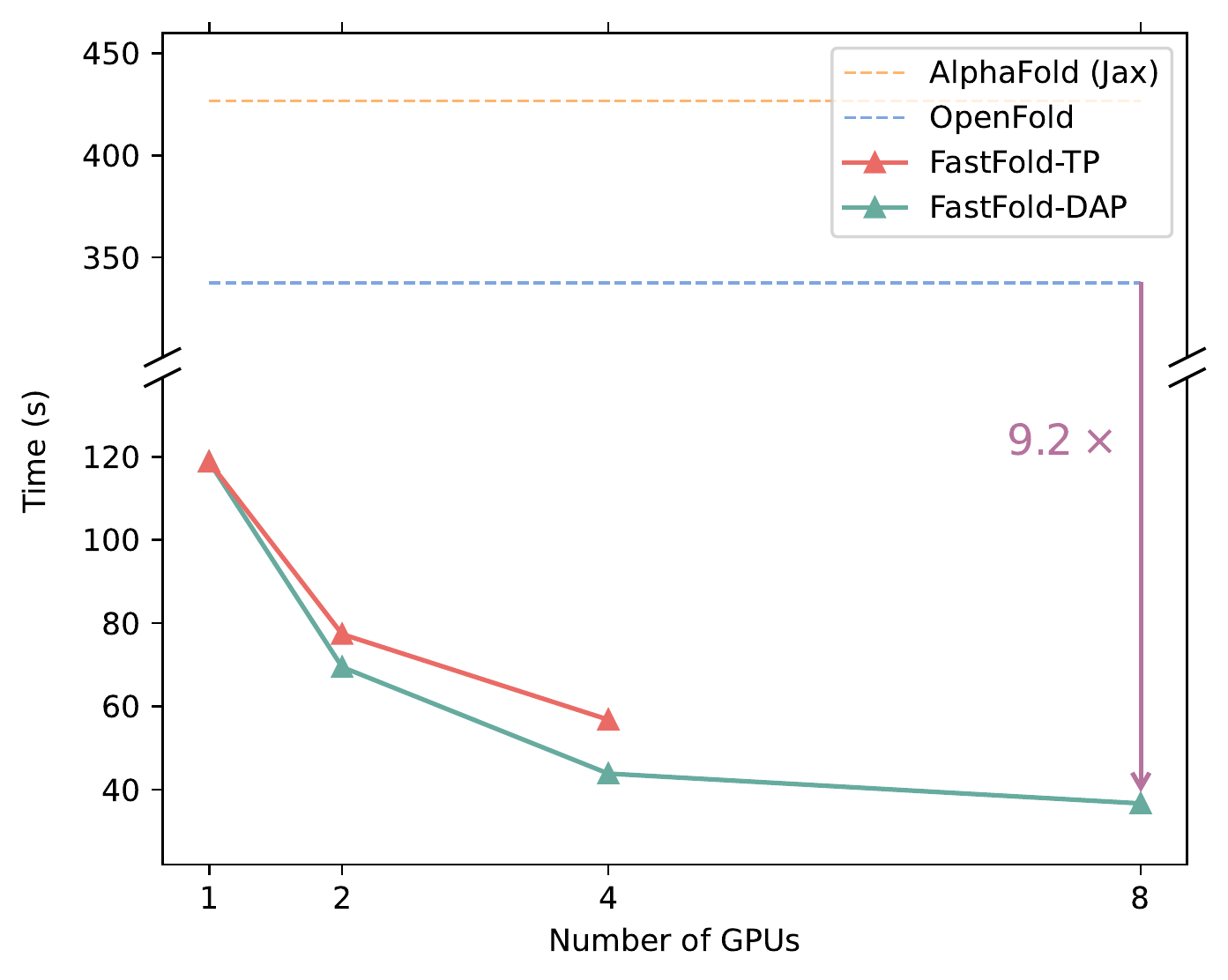}  
  }    
  \hfill
  \subfigure[Sequence Length = 2048] { 
    \centering
    \label{fig:inference_long_2048}     
    \includegraphics[width=0.60\columnwidth]{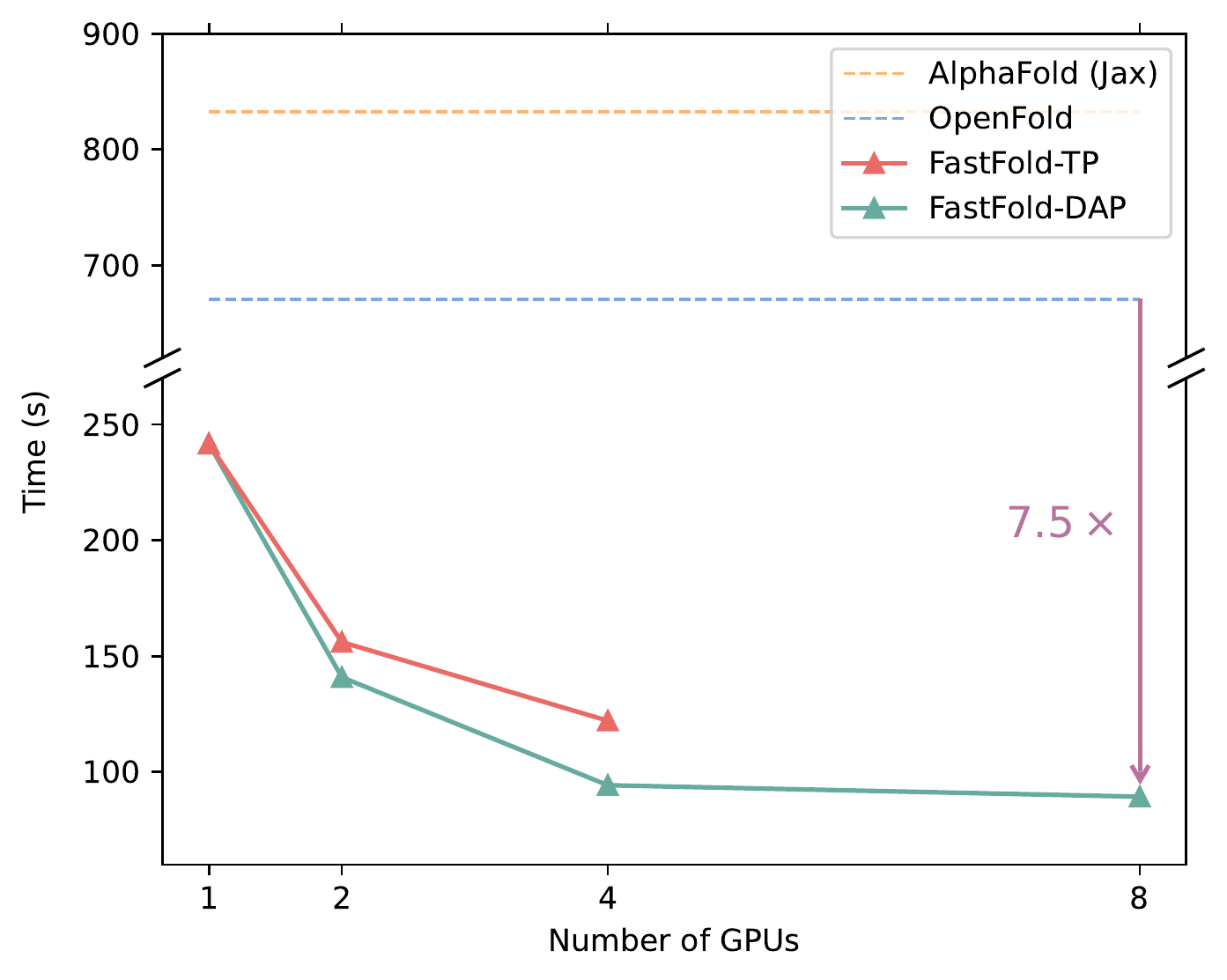}
  }
  \hfill
  \subfigure[Sequence Length = 2560] {
    \centering
    \label{fig:inference_long_2560}     
    \includegraphics[width=0.60\columnwidth]{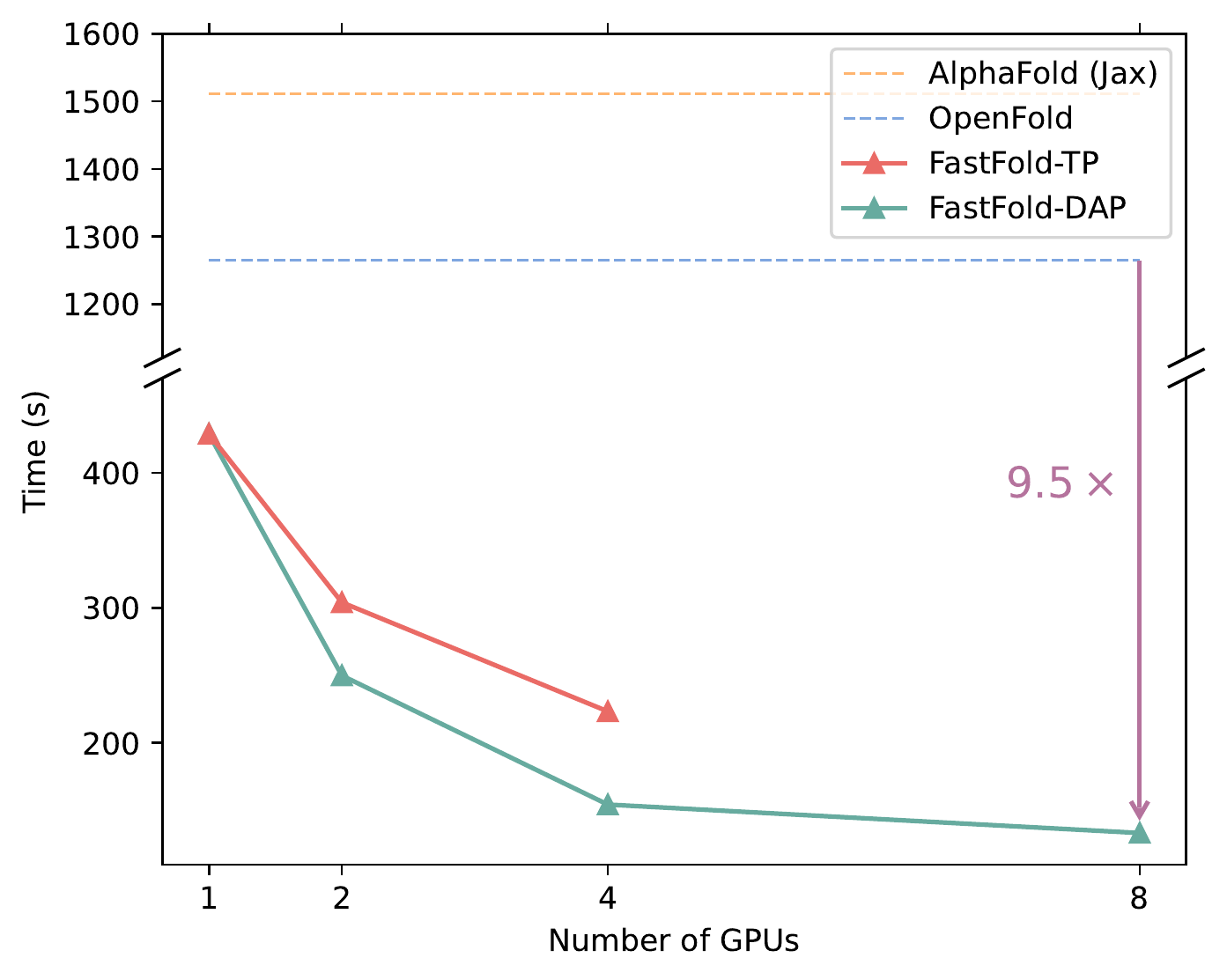}
  }
  \vspace{-5pt}
  \caption{Comparison of inference performance for long sequences. Both AlphaFold and OpenFold only support single GPU inference, so the orange and blue dashed lines are shown in the figure. \textit{FastFold-TP} and \textit{FastFold-DAP} refer to parallel inference using TP and DAP on multiple GPUs, respectively.}     
  \label{fig:inference_scale}
\end{figure*}

For long sequence inference with amino acid sequences ranging from 1K to 2.5K in length, direct inference already encounters memory capacity problems and takes several minutes or even tens of minutes. To address this issue, AlphaFold and OpenFold use the chunking technique for inference, dividing chunks along the sequence dimension and computing them sequentially in a single operation. This technique effectively reduces memory consumption during computation. In contrast, FastFold can use distributed inference to reduce the memory capacity requirement and significantly shorten the inference time. As shown in Figure \ref{fig:inference_scale}, FastFold reduces the inference time by 7.5-9.5 times compared to OpenFold and by 9.3-11.6 times compared to AlphaFold when using distributed inference. The figure also shows that DAP can scale to more GPUs (TP can only scale to 4 GPUs due to the limitations mentioned in section \ref{TP_alphafold}), and has significantly better overall scaling efficiency compared to TP.

For inference with extremely long sequences, over 3K in length, even with the chunking technique, the single GPU's memory capacity is exceeded. As shown in Table 3, both AlphaFold and OpenFold encounter Out of Memory (OOM) errors when the sequence length reaches 3K. However, FastFold can utilize distributed inference with AutoChunk for inference on extremely long sequences. In fact, for sequences up to 4K in length, FastFold's inference latency is within 10 minutes.

\begin{table}[htb]
\small
\centering
\renewcommand\arraystretch{1.2}
\caption{Inference Latency for Extremely Long Sequence (s).}
\vspace{2pt}
\begin{tabular}{ccccc}
\noalign{\hrule height 1pt}
\begin{tabular}[c]{@{}c@{}}Sequence\\ Length\end{tabular} & AlphaFold & OpenFold & \begin{tabular}[c]{@{}c@{}}FastFold \\ (4 GPU)\end{tabular} & \multicolumn{1}{l}{\begin{tabular}[c]{@{}l@{}}FastFold\\ (8 GPU)\end{tabular}} \\ \hline
2560                                                      & 1511.315  & 1265.193 & 154.422                                                   & 133.381                                                    \\
3072                                                      & OOM       & OOM      & 238.51                                                     & 201.916                                                                            \\
3584                                                      & OOM       & OOM      & 414.185                                                     & 388.691                                                                            \\
4096                                                      & OOM       & OOM      & OOM                                                     & 547.955                                                                            \\ \noalign{\hrule height 1pt}
\end{tabular}
\end{table}

FastFold can significantly reduce the inference time and economic cost for different sequence lengths and enable the prediction of protein structures for extremely long sequences. This capability of FastFold makes it well-suited for large-scale, high-throughput protein structure prediction tasks, which will greatly promote the use of protein structure prediction models in various applications.

\subsection{Evoformer Performance}

\begin{figure}[htb]
    \centering
    \includegraphics[width=0.45\textwidth]{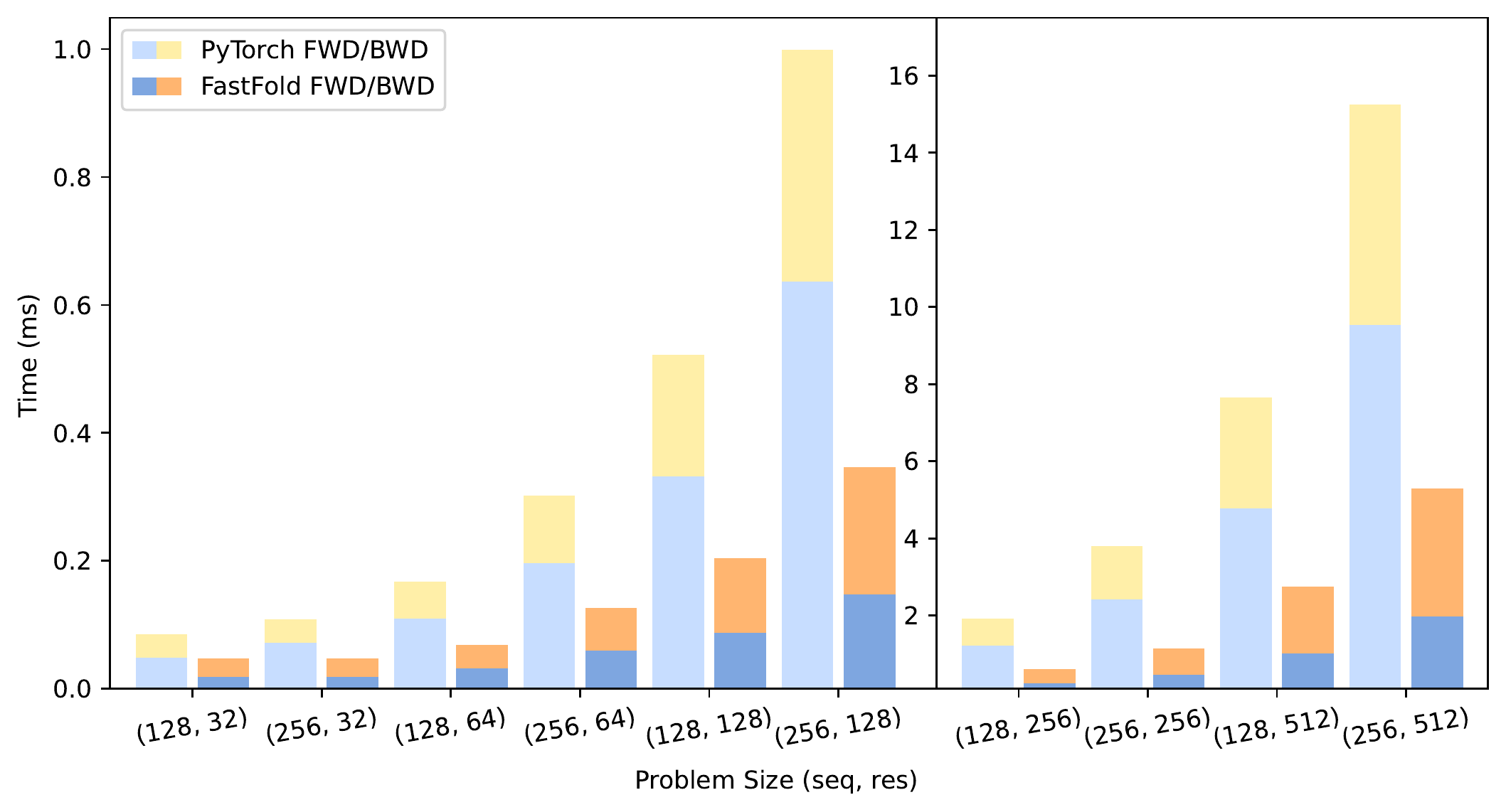}
    \vspace{-8pt}
    \caption{Performance comparison of fused softmax (BFloat16) on NVIDIA Telsa A100.}
    \label{fig:softmax_bf16}
\end{figure}

Figures \ref{fig:softmax_bf16} shows the performance comparisons of the fused softmax operations, when using Bfloat16 precision. The forward and backward latency of the PyTorch native kernel and the FastFold optimized kernel are compared. The problem size $(X, Y)$ refers to a sequence length of $X$ for the attention input and a hidden size of $Y$ for the attention. The FastFold kernel can achieve a performance improvement of $1.77\sim3.32\times$ compared to the PyTorch native kernel.


Overall, FastFold significantly improves performance of the evoformer module in both training and inference. The use of DAP and optimized kernels has allowed FastFold to scale efficiently to multiple GPUs and achieve significant speedups in both forward and backward passes. In addition, the use of the DAO has helped to reduce communication overhead and improve overall performance. FastFold has also enabled the inference of extremely long sequences, which was not possible with previous implementations. These improvements have greatly reduced the time and economic cost of training and inference, making it possible to apply the Evoformer model to large-scale protein structure prediction tasks.

\begin{table}[hbt]
\centering
\renewcommand\arraystretch{1.2}
\caption{Forward/Backward time per evoformer layer (ms).}
\label{tab:evo_speed}
\begin{tabular}{rll}
\noalign{\hrule height 1pt}
\begin{tabular}[c]{@{}r@{}}Data Size \\ (seq, res)\end{tabular} & OpenFold          & FastFold \\ \hline
(128, 256)                                                         & 17.21 / 41.76                                               & 8.31 / 16.75                                         \\
(512, 384)                                                         & 69.63 / 138.57                                             & 32.01 / 61.53                                        \\
(512, 512)                                                         & 116.02 / 233.27                                          & 54.19 / 102.83                                       \\
(1024, 384)                                                        & 135.28 / 224.11                                      & 59.41 / 152.82                                       \\ \noalign{\hrule height 1pt}
\end{tabular}
\end{table}

\subsection{AutoChunk}

\begin{figure}[htb]
    \centering
    \includegraphics[width=0.45\textwidth]{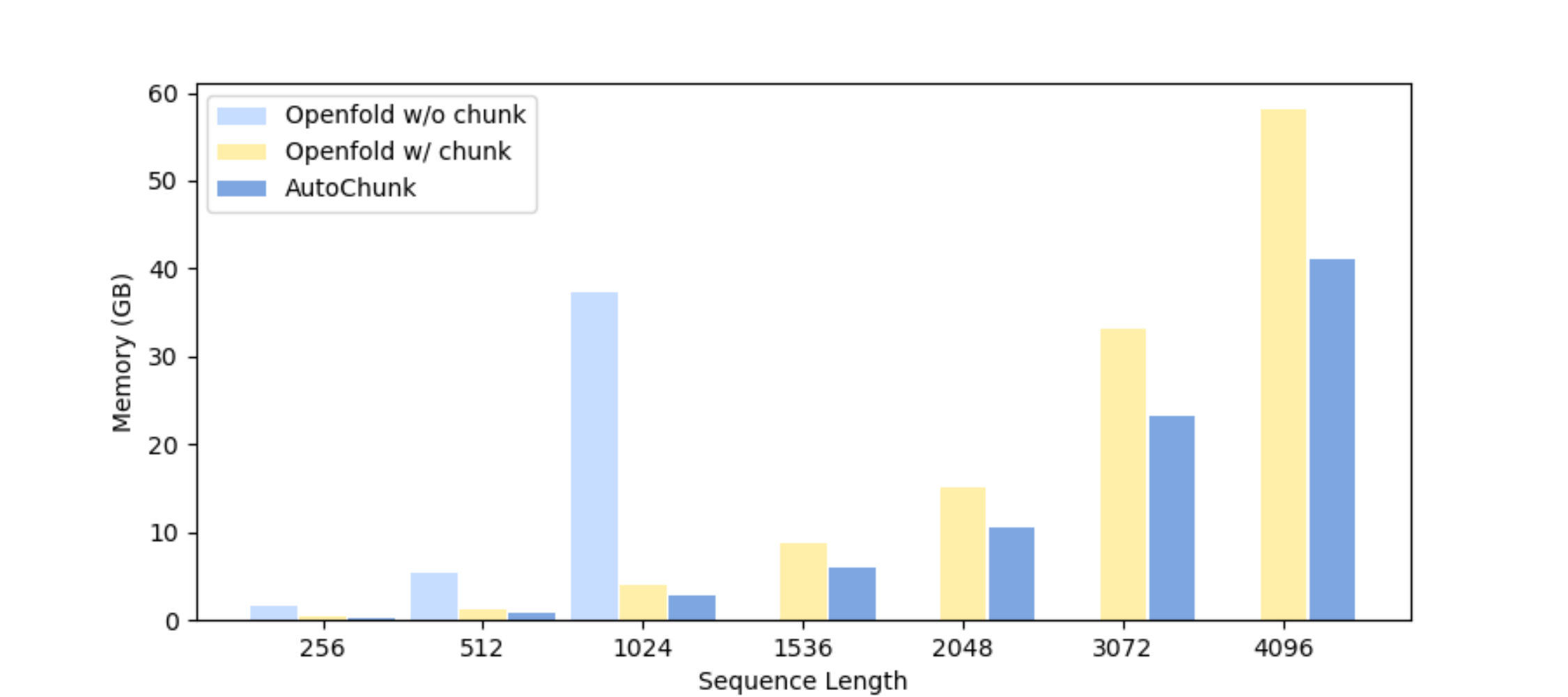}
    \vspace{-5pt}
    \caption{Memory comparison of OpenFold and FastFold AutoChunk on NVIDIA Telsa A100.}
    \label{fig:autochunk_mem}
\end{figure}

We evaluate the performance of AutoChunk in terms of memory usage and inference latency. As shown in Figure \ref{fig:autochunk_mem}, we compare the peak memory usage for the inference of various sequence lengths for Openfold without chunk, Openfold with chunk=1, and AutoChunk. When the sequence length surpasses 1024, Openfold without chunk will experience an Out of Memory (OOM) error while the other two with chunk will not. AutoChunk significantly reduces the memory usage by 86.0\%-92.6\% compared to Openfold without chunk and by 30.6\%-34.4\% compared to Openfold with chunk, which is designed by experts. This demonstrates the effectiveness of AutoChunk in significantly reducing the memory usage while also surpassing expert-designed approaches by a large margin.

\begin{figure}[htb]
    \centering
    \includegraphics[width=0.42 \textwidth]{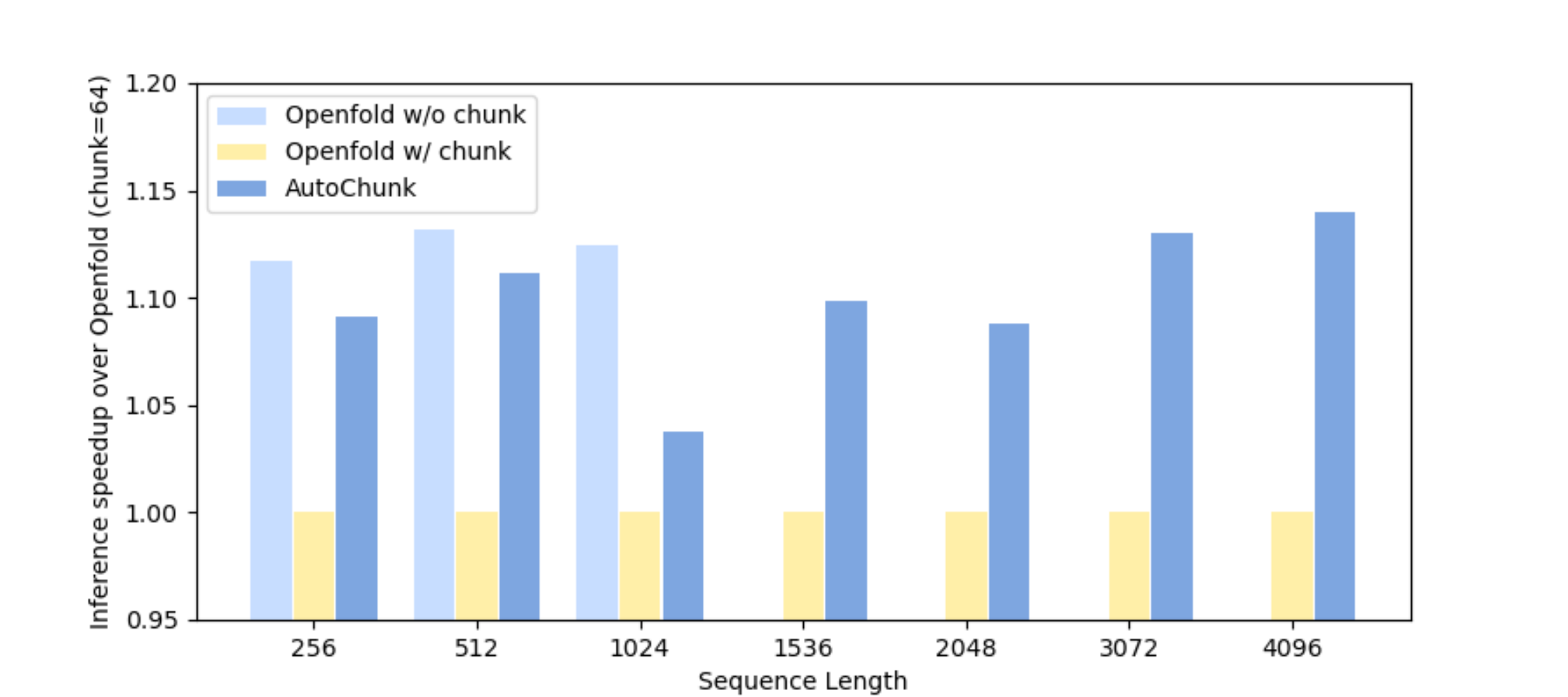}
    \vspace{-5pt}
    \caption{Inference latency comparison of OpenFold and FastFold AutoChunk on NVIDIA Telsa A100. Performance is normalized by OpenFold with chunk=64 }
    \label{fig:autochunk_infer}
\end{figure}

In Figure \ref{fig:autochunk_infer}, we compare the speedup of Openfold without chunk and AutoChunk versus Openfold with chunk when inferring sequences of different lengths. We fix the chunk size of Openfold with chunk=64, as it has been found to reduce memory usage by approximately 80\% while approaching the maximum improvement in memory, and having a relatively small impact on speed. We also set the memory limit of AutoChunk to the memory cost of Openfold with chunk=64 to ensure that the memory usage of both is the same. On average, AutoChunk improves the inference speed by 12\% compared to Openfold with chunk and only losses 4\% compared to Openfold without chunk. This indicates that AutoChunk efficiently selects the chunk range and size with minimal overhead.

\section{Related Work}

\textbf{Optimization for protein prediction models.} The optimization of protein structure prediction models for training and inference has received relatively little attention. ParaFold\cite{zhong2022parafold} is an optimized system for AlphaFold inference on heterogeneous platforms, focusing on optimizing the data processing workflow for multi-threaded parallelism on the CPU platform. FastFold, on the other hand, focuses on the training and inference of the AlphaFold model on the GPU platform. The model optimization of FastFold and the data processing workflow optimization of ParaFold are complementary and could be combined in future work.

\textbf{Efficient Transformer.} There have been many efforts to optimize the performance of Transformer models, which can be broadly classified into two categories: \textit{efficient design} of the transformer and \textit{efficient implementation} of the transformer. Many works aim to reduce the complexity of attention computation in the transformer through techniques such as sliding windows or low-rank approximation, such as Performer \cite{choromanski2020rethinking}, Reformer\cite{kitaev2020reformer}, and Linformer\cite{wang2020linformer}. Others, like LightSeq\cite{wang2020lightseq, wang2021lightseq2} and TurboTransformer\cite{fang2021turbotransformers}, focus on optimizing the inference performance of the transformer on the GPU platform through CUDA optimization and memory management. However, the evoformer has several differences from the vanilla transformer, and the optimizations in FastFold are mostly based on the characteristics of the evoformer. These optimizations can be combined with parallel techniques to significantly reduce the time costs of training and inference.

\textbf{Large scale training.} Several approaches have been proposed to address the challenges of large-scale training. Large batch training methods such as LAMB\cite{you2019large} and LARS\cite{you2017large} have been used to speed up training and address optimization issues that arise during scaling with data parallelism. For large model training, several approaches have been proposed to achieve high performance. Megatron-LM\cite{10.1145/3458817.3476209} uses a hybrid parallel strategy to scale the model to more GPUs. DeepSpeed ZeRO\cite{rajbhandari2020zero, 10.1145/3458817.3476205} provides a memory-efficient optimizer. In FastFold, we proposed \textit{Dynamic Axial Parallelism}, which has higher scaling efficiency than current mainstream model parallelism methods.

\section{Conclusion}

Accurate prediction of protein structure is essential for progress in structural biology, and devising efficient methods for training and inference these models is a significant challenge. FastFold aims to tackle this challenge by utilizing Dynamic Axial Parallelism, which enables both training and inference to be executed across multiple GPUs, resulting in a considerable decrease in time consumption. Furthermore, FastFold incorporates various low-level optimization techniques, such as AutoChunk and kernel optimization, to further enhance efficiency and minimize the cost of training and inference. These methods allow for the design and deployment of more efficient protein structure prediction models and also facilitate the creation of larger models for improved performance.

The optimization technique presented in this work is versatile and applicable to a wide range of models. As most protein prediction models employ similar structures, the optimization technique can be easily implemented on models such as RoseTTAFold\cite{baek2021accurate}, MSA Transformer\cite{rao2021msa}, and others. Analogously, the video transformer uses axial attention along the spatial dimension or temporal dimension respectively. Our proposed communication optimization techniques can be tailored to video models\cite{arnab2021vivit, ho2019axial}.

\bibliographystyle{plain}
\bibliography{tex/sample-base}

\end{document}